# TriView-YOLO: Early Multi-View Fusion for Ground Penetrating Radar Cavity Detection in Soft, High-Water-Content Soils

Suphawut Thawinutchokaudom[1], Sompote Youwai[1*], Warat Kongkitkul[1], Mitsumasa Yamashina[2], José M.D.S. Rodrigues Neto[3], Jun Shinohara[2]

[1] *AI Research Group, Department of Civil Engineering, Faculty of Engineering, King Mongkut's University of Technology Thonburi, Bangkok, Thailand*

[2] *Center for Disaster Management Informatics Research, Ehime University, Matsuyama, Japan*

[3] *Canaan Geo Research Co., Ltd., Matsuyama, Ehime, Japan*

[*]Corresponding author: sompote.you@kmutt.ac.th

***Abstract*—Automated detection of subsurface cavities from Ground Penetrating Radar (GPR) is most difficult in soft, high-water-content ground, where conductive, water-saturated soil attenuates the signal and degrades cavity reflections, yet this is also the condition under which cavities most readily form. This paper proposes TriView-YOLO, a multi-view YOLOv12 detector for road cavity screening in such ground. Three co-registered views (longitudinal B-scan, horizontal C-scan, and cross-section B-scan) form a 9-channel input fused by a TripleInputConv layer that replaces the YOLOv12 stem; the rest of the network is unchanged, and bounding boxes are required on the longitudinal view only. Training used 1,600 expert-verified field samples, principally metropolitan road surveys of Bangkok, Thailand, acquired with a vehicle-mounted multichannel three-dimensional GPR mobile mapping system, with surveys over the firmer subgrades of Japan added to training and validation only. The test set comes exclusively from the Bangkok surveys, over soft marine clay with 80–140% water content and a water table at 1–2 m depth, a ground condition for which no dedicated deep learning cavity-detection evaluation has been reported. On this unaugmented, field-only test set, split randomly within surveys, the proposed model attains mAP50 0.558 ± 0.028 over three seeds at 23.6 GFLOPs and 3.1 ms per image. Ablations show that removing the auxiliary views lowers mAP50 and recall, whereas public and synthetic training images, DINOv3 features, larger model scale, and COCO pretraining bring no gain.**

## I. INTRODUCTION

Cavities that form beneath road pavements are a latent hazard to the travelling public. They grow silently from groundwater flow, soil erosion, utility leakage, inadequate compaction, and the deterioration of buried structures, and they fail without warning, so that the first visible symptom is often collapse of the carriageway under live traffic [1]; in Bangkok, repeated sinkhole incidents on major corridors have prompted agency-level investigation of high-risk sections [2]. Because early-stage cavities give no reliable surface signature, they escape visual inspection, and an authority acting only on observable distress acts after the safety margin is lost.

Preventive screening at network scale is therefore a transportation-safety function, and vehicle-mounted sensing has made it practicable. Ground penetrating radar (GPR) records reflections generated by contrasts in dielectric permittivity [3], air-filled voids produce strong boundary reflections, and a multichannel array with satellite positioning can cover hundreds of lane-kilometres of a city network in one campaign [2]. The bottleneck has moved to interpretation, which remains difficult where overlapping reflections and low signal-to-noise ratios occur [4] and which takes weeks of analyst-dependent manual reading at city scale [5]. Learned detectors have accordingly been brought to pavement GPR, including YOLO variants for concealed cracks [6] and asphalt voids [7], transformer classifiers of pavement condition [8], and volumetric interpretation from 3D arrays [9].

The task is most demanding in soft ground with high water content. Deltaic cities such as Bangkok overlie 8–15 m of very soft marine clay with natural water content of 80–140% and a groundwater table at only 1–2 m depth [10]. The hydrogeological conditions that favor cavity formation, namely shallow groundwater and water-saturated erodible soils around leaking utilities, also degrade the radar signal required to locate cavities. Wet, conductive clay strongly attenuates electromagnetic waves, so cavity reflections are weaker, depth-limited, and noisier than in the firm or well-drained soils in which nearly all published GPR cavity-detection studies have been conducted. Automated detection in this environment therefore cannot be assumed to follow from methods validated on firm ground and must be demonstrated directly.

Among detection frameworks, the You Only Look Once (YOLO) family combines accuracy with the throughput that survey-scale screening requires. The recent YOLOv12 introduces attention-centric components, namely Area Attention and R-ELAN blocks, that improve detection of small, irregular, and partially occluded objects while preserving inference speed [11], characteristics that suit GPR imagery in which cavity reflections are weak and partially obscured.

This study differs from previous research in three respects. Regarding the survey environment, the field datasets behind published cavity detectors were collected on firm or well-drained ground, in Seoul [12], [13], Shenzhen [14], Chinese highways [15], [16], and airport runways [17], whereas the present study evaluates exclusively on Bangkok subsoil surveys over soft marine clay in which GPR interpretation has so far been manual [18]. Regarding the input representation, nearly all previous detectors operate on a single B-scan. Where multiple views have been exploited, they have served image-level classification from 3D GPR volumes [9] or semantic segmentation of pavement objects from several perspectives [19], while the one prior multi-view detector, MV-GPRNet [17], requires raw volumetric data from a dedicated 3D array and a custom volumetric fusion network. The present study instead fuses three co-registered views through a lightweight input-level modification of a standard detector operating on ordinary images exported from commercial GPR software, and keeps bounding-box annotation on one view. Regarding the evaluation protocol, every configuration is trained under three random seeds and reported as mean ± standard deviation, with accuracy analyzed jointly with FLOPs and latency. Because every accuracy is measured on unaugmented field data from a high-attenuation soft-clay environment, the absolute mAP50 values are not directly comparable with the 0.87–0.96 figures reported under single-source or augmentation-expanded evaluations (Section V-B).

The objectives are: (1) to develop TriView-YOLO, combining YOLOv12 with a TripleInputConv stem that fuses three co-registered GPR views into a 9-channel input while requiring labels on one view only; (2) to quantify automated cavity screening in a soft-ground, high-water-content environment under a field-only protocol, so that road authorities can calibrate expectations against a realistic figure rather than against accuracies from favourable ground; and (3) to isolate the contribution of each design element through seeded ablations. Throughout, the detector is treated as a screening aid whose operating point is set by the risk asymmetry of the application: a missed cavity is a latent safety hazard, whereas a false alarm costs an expert review, so recall is weighted above precision and every flagged location is passed to human verification.

## II. RELATED WORK

Subsurface cavities develop through initiation, enlargement, and critical instability driven by structural deterioration, traffic loading, and groundwater variation. Deep learning detectors have been applied along three lines: single-view detection on B-scan images with successive YOLO generations and domain-specific modifications [20], [21], [6], [22], [14], [8], [7], [15], [16], [23], [24], [25]; 3D or multi-view approaches exploiting volumetric data, for image-level classification [12], [13], [9], for multi-perspective semantic segmentation [19], for detection with custom volumetric fusion networks on dedicated 3D arrays [17], or for post-hoc cross-verification of single-view detections [26]; and data-scarcity strategies using simulation, public datasets, or pretrained features [27], [28], [29]. Table I positions representative studies against the present work. Accuracies obtained on controlled or single-source datasets (mAP ≈ 0.87–0.96) are substantially higher than those reported for heterogeneous field data with expert-derived labels (Section V-B).

TABLE I
REPRESENTATIVE DEEP LEARNING STUDIES FOR GPR-BASED SUBSURFACE DEFECT DETECTION, COMPARED WITH THIS STUDY.

| Study | Base model | GPR input | Data | Reported performance | Limitation relative to this study |
|---|---|---|---|---|---|
| Kang et al. [12], [13] | CNN (UcNet) | 3D GPR image grids | Seoul urban roads | Improved discrimination | Image-level classification; no localization |
| Li et al. [17], MV-GPRNet | Custom 3D CNN fusion | 3D C-scan volume + top views | Airport runway 3D array | F1 (void) 91% | Requires dedicated 3D array and custom volumetric network |
| Yang et al. [22], CP-YOLOX | Improved YOLOX | Single view from 3D GPR | 209.6-km highway, Jiangxi | mAP 0.806; recall 0.587 | Single view; recall < 0.6 on field-only test |
| Kan et al. [14] | MHUnet + YOLOv8 | Enhanced single view | ~1,700 images, Shenzhen | mAP 0.876 | Detection stage still single-view |
| Chai et al. [16], ACF-YOLO | Modified YOLOv8 | Single B-scan | FKS-GPR (350→2,800 images) | mAP 92.2% | Heavy augmentation; single view |
| Gao et al. [23] | YOLOv5 | Single B-scan | 1,248 simulated + field images | Accuracy 0.911 (voids) | Synthetic-dominated training and evaluation |
| Peng et al. [26] | YOLO + cross-view verification | Views verified post hoc | 1,250 km urban 3D GPR (field-only) | Void precision 0.324, recall 0.318 | Views cross-checked after detection, not fused as input |
| Zhang et al. [25], FLS-YOLO | Lightweight modified YOLO | Single B-scan | 3,587 field + 1,000 synthetic | mAP50 0.696 in-domain; 0.558 zero-shot | Single view; synthetic scans in test split |
| **This study (TriView-YOLO)** | **YOLOv12 + TripleInputConv (9-channel)** | **Three co-registered views** | **1,600 field samples; Bangkok-only test set** | **mAP50 0.558 ± 0.028** | — |

A separate body of literature quantifies how severely soft, wet ground constrains GPR detection of any target. Attenuation reaches 1–300 dB/m in clays, against 0.01 dB/m in dry sand at 100 MHz [30]. Blind discrimination tests across soil types show detection performance degrading several-fold in moisture-affected soils [31]. For the Bangkok subsoil specifically, the only published GPR void study obtained usable penetration below 2 m with a 400 MHz antenna, interpreted radargrams manually, and required drilling to confirm the void [18]. Geophysical evidence thus establishes soft, high-water-content ground as the most adverse detection environment, yet no published study reports a cavity detector both developed and evaluated for such a deltaic environment. The present study addresses this gap by combining multi-view evidence fusion with field-only evaluation in this ground condition.

## III. METHODOLOGY

### *A. Overview*

Cavity detection was formulated as single-class object detection (“cavity”), with images containing no cavity retained as background samples. The proposed model, TriView-YOLO, is a multi-view YOLOv12 detector in which longitudinal B-scan, horizontal C-scan, and cross-section B-scan images from the same location form a joint 9-channel input fused by a TripleInputConv stem (Section III-D). An ablation study (Section III-E) evaluates single-view detection, multi-source training data, DINOv3 feature enhancement, model scale, and initialization under identical conditions. Each configuration was trained with three random seeds (0, 1, 2) under fixed data splits, and results are reported as mean ± standard deviation.

### *B. Data acquisition*

Four data sources were used, in three separated roles: Bangkok field data, which are the target environment and the only source of test images; Japan field data, which add training and validation images from a contrasting subgrade; and Roboflow Universe and gprMax images, which enter the training set of one ablation only. The two field sources alone train and evaluate the proposed multi-view model.

Bangkok roads overlie the very soft, high-water-content marine clay of the Chao Phraya delta (8–15 m of soft clay, water table at 1–2 m; [10]), in which high electrical conductivity strongly attenuates GPR signals. The Bangkok images were acquired with the same system used in a metropolitan-scale road survey programme in Bangkok, the Ground Penetrating Radar Mobile Mapping System 3D, a vehicle-mounted system that couples a multichannel stepped-frequency array of 20 antennas (100–3000 MHz, effective swath about 1.5 m) with GNSS and omnidirectional-camera mobile mapping, and that characterizes subsurface features to typical depths of 1.5–2.0 m in Bangkok conditions [2]. In that programme, campaigns in February 2025, January 2026, and March 2026 covered 823.1 lane-km of Bangkok roads and identified 616 potential cavities, an overall potential cavity rate of 0.75 per lane-km, with markedly higher rates in curb-side external lanes than in internal lanes (1.09 against 0.40 per lane-km on urban main roads) and six high-risk cavities verified by borehole camera inspection. Major Bangkok roads commonly use composite pavement, an asphalt overlay on a concrete slab, so the radargrams combine soft-clay attenuation with the clutter and reduced penetration of concrete-dominant structures. B-scan and C-scan radargram images were reviewed and exported from road pavement survey records managed in the Kontur Examiner system and were treated as image inputs rather than as raw electromagnetic signals. These surveys supply the co-registered three-view samples and, exclusively, the test split of every dataset (Section III-C).

Survey routes in Japan were exported in the same image form from the same survey-record system and represent firmer, better-drained subgrades. They form a small auxiliary fraction of the training data (8% of the multi-view training samples; Section III-C), included so that the training set contains low- as well as high-attenuation reflection characteristics, and they are excluded from the test split; no accuracy reported here is measured on Japanese images. Two further sources were collected solely to test whether scarce field data can usefully be supplemented with cleaner external images. Roboflow Universe public data comprised publicly available GPR images screened for relevance to cavity detection. gprMax synthetic data were B-scans generated by finite-difference time-domain simulation [27] of a three-layer pavement model (0.10 m asphalt over 0.20 m concrete over soil), with fixed domain height 1.5 m, dx = 0.01 m, 20 ns time window, and 150 traces at 0.05 m spacing; antenna frequency (200/300/500 MHz), soil permittivity (5–15), conductivity (0.001–0.02 S/m), target radius (0.05–0.2 m), depth (0.2–1.2 m), and object type (air void, metal, none) were varied. Neither source provides the three co-registered views that the multi-view input requires, so both are used only in the single-view multi-source configurations of the ablation study, and only within their training split; the proposed multi-view model is never trained on them.

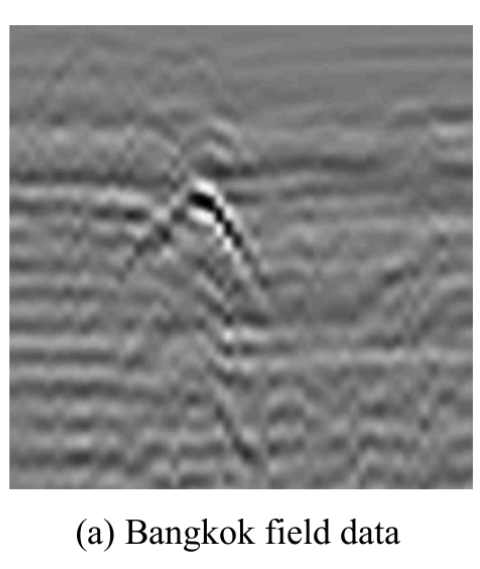

(a) Bangkok field data

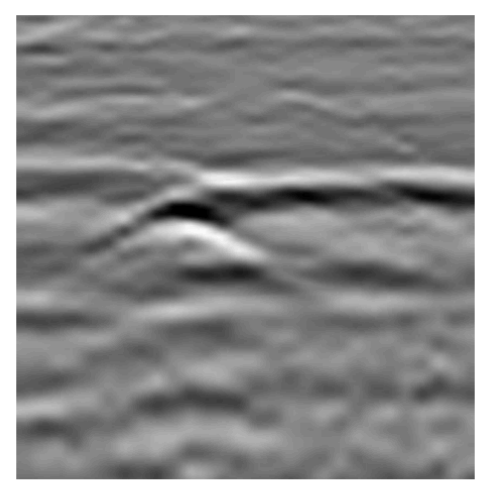

(b) Japan field data

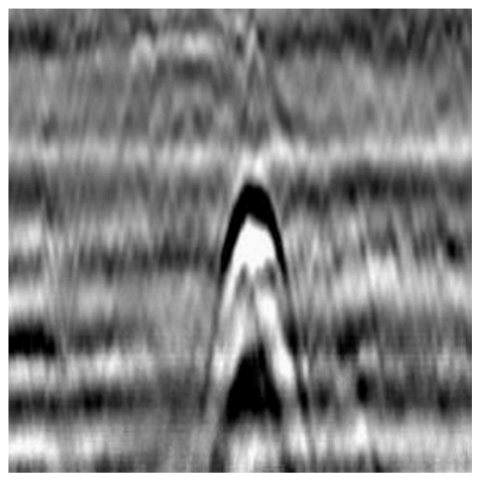

(c) Roboflow Universe data

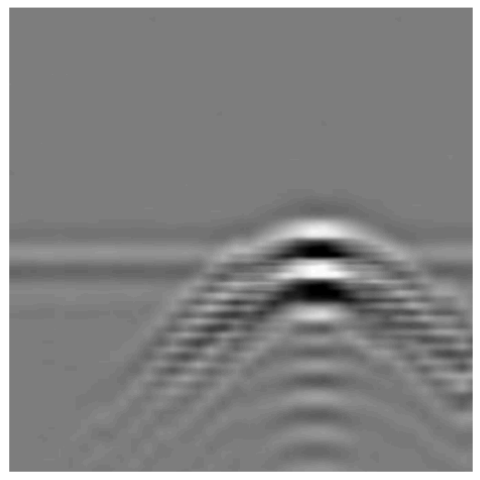

(d) gprMax synthetic data

Fig. 1. Longitudinal B-scan examples from the four data sources: (a) Bangkok field data (soft, high-water-content marine clay), (b) Japan field data (firmer subgrade), (c) Roboflow Universe public data, and (d) gprMax synthetic data.

The contrast between the two field panels of Fig. 1 reflects ground conditions rather than acquisition equipment. In the Bangkok example the cavity hyperbola is faint, broken, and embedded in horizontal ringing bands and speckled clutter. Free water has a relative permittivity of approximately 80 against 3–10 for dry pavement materials, so the water content of the Bangkok clay, with its high conductivity, causes losses of 1–300 dB/m against approximately 0.01 dB/m in dry sand [3], [30]; the shallow water table generates high-amplitude ringing that overprints the target signature, and heterogeneous moisture fills the background with incoherent clutter [4]. The Japan example, over a firmer and better-drained subgrade, shows a smoother background in which the cavity reflection remains a coherent hyperbola.

### *C. Image preparation and ground truth*

Reference cavity locations were provided by GPR experts from prior survey interpretation. Before cropping, each location was cross-checked using three interpretation indicators: a hyperbolic or cavity-related reflection in the B-scan, a dark localized pit-like anomaly in the horizontal C-scan, and a clear amplitude variation in the corresponding A-scan trace [32]. Because hyperbolic reflections can also arise from pipes, moisture, metallic objects, or layer interfaces, these indicators were used jointly as consistency checks rather than standalone criteria. Images were cropped manually over 4–6 m distance intervals along the survey line, long enough to contain the cavity signature and its background without excessive rescaling to the 640 × 640 network input. For multi-view samples, the horizontal C-scan and the cross-section B-scan were cropped over the same distance interval and a consistent depth range as the longitudinal B-scan; the depth of the exported C-scan slice follows the slice at which the anomaly was localized during the joint expert interpretation, and is therefore an interpretation-dependent choice rather than a fixed acquisition parameter.

Cavity identification and labeling were carried out by an expert team with more than ten years of experience in subsurface cavity surveying in Japan, and a subset was verified by drilling, boreholes confirming both the existence of the cavity and its size. Annotation was performed in Roboflow. Bounding boxes were drawn only on longitudinal B-scans, which are the labeled view in both the single-view and multi-view settings, enclosing the hyperbolic signature across bright and dark bands with the box center near the apex. Images without cavities were retained as unannotated background samples. Because expert-identified positions in attenuating ground rest on degraded signatures, all labels were re-checked under the same three-indicator protocol before the final experiments: box extents were corrected, overlooked faint signatures were annotated, and boxes not supported by the joint evidence were removed. The resulting annotation set, whose test split contains 133 cavity instances, is common to every configuration reported in Section IV, so all rows of Table II are compared at identical labels and data. Because the re-check followed an initial training cycle, blind re-annotation by independent interpreters with inter-rater agreement statistics is identified as future work.

**Dataset splits (80:10:10).** The field single-view dataset comprised 1,280 / 160 / 160 images. The multi-source single-view dataset comprised 1,452 / 182 / 182 images, with Roboflow and gprMax images added to the training set only, so that validation and test sets contained field images exclusively. The multi-view dataset comprised 1,280 / 160 / 160 samples, each consisting of three matched views (4,800 view images in total); of the 1,280 training samples, 1,183 (92%) come from the Bangkok surveys and 97 (8%) from the Japanese routes, so the training distribution is dominated by the target environment. In every dataset the test split was constrained to images from the Bangkok subsoil surveys, so that the reported accuracy refers to a single, adverse ground condition, whereas training and validation splits contain both Bangkok and Japanese images. Subject to that constraint, the splits used a fixed random assignment shared by all configurations and seeds. Images cropped from nearby sections of the same Bangkok route can therefore fall into different splits, so the reported accuracies measure within-survey generalization and may be optimistic relative to deployment on unseen routes; evaluation on fully independent routes is the principal limitation of the protocol (Section V-C).

### *D. Proposed architecture: TriView-YOLO*

TriView-YOLO is a YOLOv12-small detector whose input stage is replaced by a **TripleInputConv** layer that fuses the three co-registered views at the earliest point in the network. This configuration is denoted **MV small** in the results; all alternatives in Section III-E are ablations of it. Three design principles determined the architecture. First, fusion is applied at the stage where the data are aligned: the multi-modal literature reports that the optimal fusion stage depends on how well the modalities are registered and how far their statistics diverge [33], and the three GPR views are co-registered by construction and share the same grayscale radargram statistics, which is the regime in which input-level fusion is adequate. Fusion at the first layer also allows the auxiliary views to influence every subsequent feature, in contrast to post-hoc cross-view verification, which can only filter detections already made from a single view [26]. Second, the modification is confined to the stem, which preserves compatibility with pretrained weights and ensures that any difference from the single-view counterpart is attributable to the input representation alone. Third, the cost remains consistent with screening use: multi-view alternatives require a dedicated 3D array and a volumetric fusion network [17] or dual backbones with per-modality streams [33], whereas the stem-level design adds only two first-layer branches and a 1 × 1 fusion to a single shared backbone (23.6 versus 21.2 GFLOPs) and requires annotation on one view.

Although radargrams are grayscale, each view is loaded in three-channel format, so that each branch remains a direct copy of the standard stem and can inherit pretrained first-layer filters unchanged. Let $x^{(L)}, x^{(C)}, x^{(S)} \in \mathbb{R}^{3\times H\times W}$ denote the longitudinal B-scan, horizontal C-scan, and cross-section B-scan of the same location, with image height and width $H = W = 640$. Writing $[\,\cdot\,;\ \cdot\,]$ for concatenation along the channel dimension, the network input is

$$X = \left[\, x^{(L)}; x^{(C)}; x^{(S)} \,\right] \in \mathbb{R}^{9\times H\times W}, \tag{1}$$

so that channels 0–2 hold the longitudinal B-scan, channels 3–5 the C-scan, and channels 6–8 the cross-section. Only the longitudinal view carries bounding-box labels; the remaining views act as supplementary spatial evidence. Each location contributes one image file per view under a common name, so the views are paired automatically at load time; each auxiliary view is resized to the primary image before concatenation, and a missing auxiliary view is replaced by a copy of the primary image so that the 9-channel format remains valid. The views are ordinary images exported by the survey software, so the detector requires neither access to raw volumetric data nor a custom volumetric network, although a multichannel acquisition system remains necessary to produce the C-scan and cross-section views.

In standard YOLOv12 the first backbone layer is a single strided convolution $\mathrm{Conv}(3, c_2, k=3, s=2)$, where $c_2$ is the output channel width of the stem ($c_2 = 32$ at the small scale used here). TripleInputConv replaces this layer with a branch-then-fuse design: the input $X$ is split back into its three views, each view $v \in \{L, C, S\}$ is processed by an independent branch of convolution, batch normalization BN, and SiLU activation $\phi(z) = z\,\sigma(z)$, in which $\sigma$ is the logistic sigmoid, and the branch outputs are concatenated and projected back to $c_2$ channels by a pointwise convolution,

$$\begin{aligned} F^{(v)} &= \phi\Big(\mathrm{BN}(W^{(v)} *_2 x^{(v)})\Big) \in \mathbb{R}^{c_2 \times (H/2) \times (W/2)}, \\ F_0 &= \phi\Big(\mathrm{BN}(W^{(f)} *_1 [\,F^{(L)}; F^{(C)}; F^{(S)}\,])\Big), \end{aligned} \tag{2}$$

where $*_s$ denotes convolution with stride $s$, $F^{(v)}$ is the branch feature map of view $v$ and $F_0$ the fused stem output, $W^{(v)} \in \mathbb{R}^{c_2 \times 3 \times 3 \times 3}$ is the branch kernel of view $v$, and $W^{(f)} \in \mathbb{R}^{c_2 \times 3c_2 \times 1 \times 1}$ the fusion kernel. Each view is thus assigned an independent set of low-level filters, which is necessary because the views have different geometric meanings: the two B-scans are depth–distance sections in which cavities appear as hyperbolic reflections, whereas the C-scan is a plan-view amplitude map at a fixed depth, so a shared filter bank would be forced to compromise between incompatible signal statistics. The $1 \times 1$ kernel then learns a cross-view weighting at every spatial position, so that C-scan evidence can reinforce a weak longitudinal hyperbola at the same location. Relative to the standard stem, whose parameter count is $27c_2$, the additional cost is two further $27c_2$ branch kernels and the $3c_2^2$ fusion kernel; the added parameters are negligible, whereas the measured compute rises from 21.2 to 23.6 GFLOPs because the stem operates at half the input resolution.

In architectural terms the branch stage is a grouped convolution with three groups over the 9-channel input [34], and the branch-then-fuse pattern is a grouped-then-pointwise factorization of the kind used in aggregated-transformation designs [35]; the layer is therefore not claimed as a new operator. The contribution is the application-level finding and its supporting protocol: input-level grouped fusion suffices for co-registered GPR views, the pretrained first layer of a single-view detector can be replicated into every branch, and annotation is required on one view only. To reuse COCO-pretrained weights despite the modified input stage, the first-layer weights of the pretrained single-view model were replicated into all three branches together with their batch-normalization parameters and running statistics, and the fusion kernel was randomly initialized.

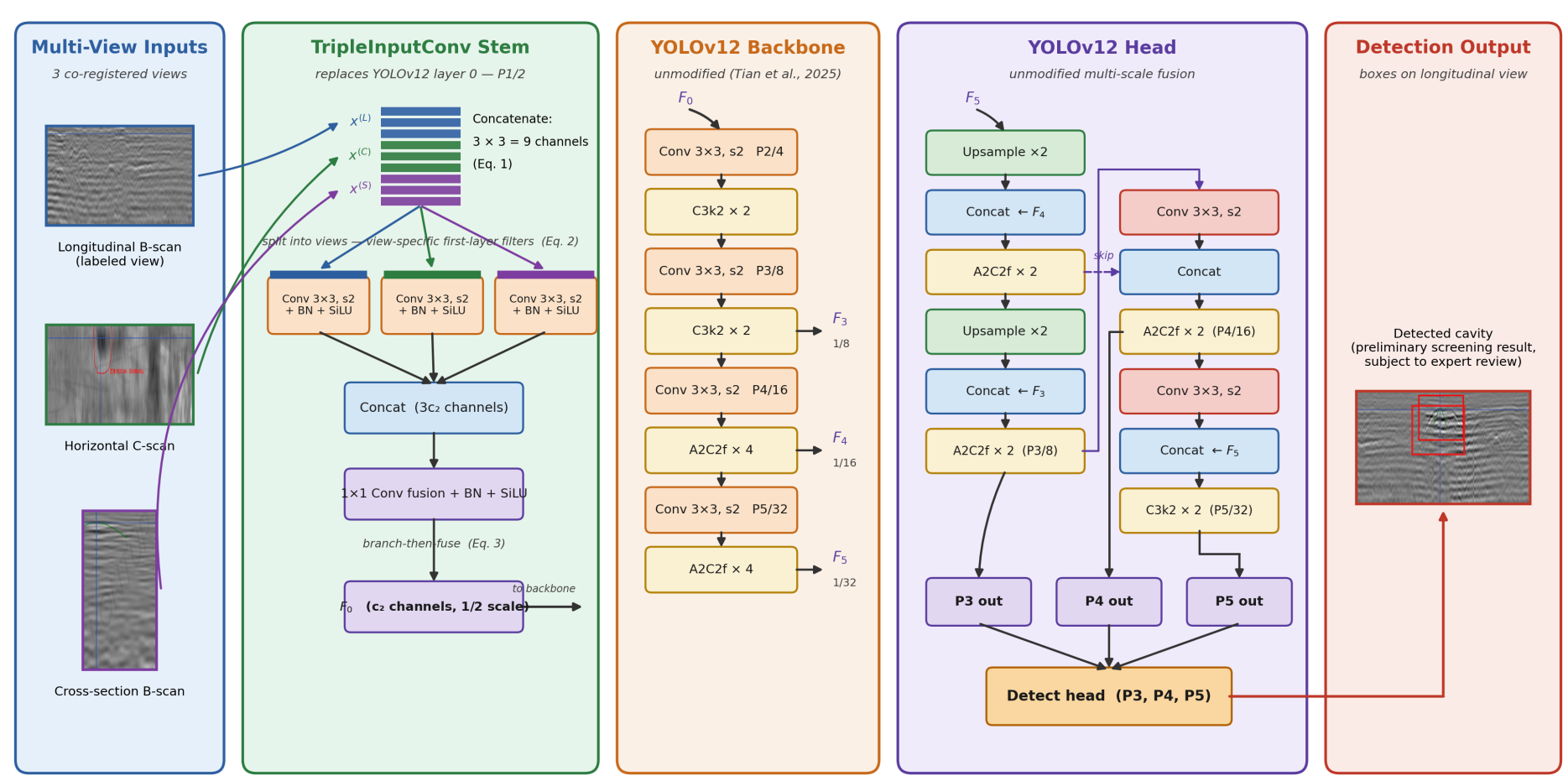

Fig. 2. Proposed multi-view YOLOv12 architecture. The three co-registered views are concatenated into a 9-channel tensor (Eq. 1) and fused by the TripleInputConv stem, which applies a view-specific stride-2 convolution branch to each view and merges them by concatenation and $1 \times 1$ convolution into $F_0$ (Eq. 2). $F_0$ enters the unmodified YOLOv12 backbone, whose $F_3/F_4/F_5$ taps feed the unmodified head and the P3/P4/P5 Detect head; predicted boxes refer to the longitudinal view.

The fused stem output $F_0$ enters the unmodified YOLOv12 backbone and head [11], which alternate stride-2 downsampling convolutions with C3k2 blocks at the shallow P2/4 and P3/8 stages and attention-centric A2C2f blocks at the deeper P4/16 and P5/32 stages, the notation P$n/s$ denoting the pyramid level and its stride. The backbone exposes the feature pyramid $\{F_3, F_4, F_5\}$ at strides 8, 16, and 32, which the unmodified head fuses along the standard top-down and bottom-up paths before the P3/P4/P5 Detect head. Within the A2C2f blocks, Area Attention partitions each feature map into $l$ non-overlapping areas ($l = 4$ by default) and computes scaled dot-product attention within each area, reducing the quadratic attention cost by a factor of $l$ while retaining a large receptive field. Retaining the backbone and head without modification serves three purposes: the attention-centric components are well matched to weak, clutter-obscured cavity reflections; an unchanged downstream network ensures that the multi-view gain is attributable to the input representation alone; and the head requires no multi-view awareness, because the fusion of Eq. (2) merges the views before the features reach it. The small (s) scale was adopted because the field dataset is too small to support larger capacity, an expectation confirmed by the scale ablation, yielding 23.6 GFLOPs and 3.1 ms single-image latency.

### *E. Ablation study design*

Each element of the proposed model was ablated under identical data splits, seeds, and metrics. Five factors were tested: input views (the proposed 9-channel multi-view input versus single-view detection on the longitudinal B-scan only, denoted 2D); training data (field images only versus field images combined with Roboflow and gprMax images, with field-only validation and test sets in both cases; tested for the single-view configurations, since the auxiliary sources provide no co-registered views); DINOv3 enhancement (none versus injection at the P4 or P5 level); model scale (small versus nano and medium); and initialization (random versus COCO-pretrained). For the DINOv3 ablations, a frozen DINOv3 vision transformer [29] provides an auxiliary feature stream, adapting the DINO-YOLO integration concept, which improved data-efficient detection on civil engineering imagery [28], to the multi-view GPR setting. The 9-channel input is reduced to three channels by a pointwise convolution, resized to 224 × 224, and encoded by the frozen transformer (ViT-S, width 384); the resulting patch tokens are projected to the channel width of the target pyramid level, fused with the backbone feature by a pointwise adapter and channel and spatial attention gates, and added residually. Only the adapter, projection, and attention parameters are trainable.

### *F. Evaluation protocol*

Detection accuracy was evaluated using precision, recall, F1-score, and mean average precision at an IoU threshold of 0.5 (mAP50) [36], reported for the single cavity class; stricter multi-threshold metrics such as mAP50–95 are not analyzed here. Computational efficiency was quantified by GFLOPs and single-image inference latency. Training stability across seeds was summarized by the coefficient of variation of mAP50, CV(%) = (SD/mean) × 100, used as a descriptive heuristic. Two statistical caveats govern the interpretation of the results: three seeds are too few for formal hypothesis testing, so differences are reported descriptively and interpreted only where the direction of the effect is consistent across seeds; and eighteen configurations are compared on a single test set, so the identity of the best configuration is subject to selection effects.

All models were implemented and trained in the Ultralytics framework (v8.3.63) under identical hyperparameters. Each model was trained for up to 150 epochs at 640 × 640 resolution with batch size 16 (nominal batch 64 via gradient accumulation) and early stopping at patience 60, and the checkpoint with the best validation fitness was retained. Optimization used AdamW (initial learning rate $1.5 \times 10^{-4}$ decayed to 1% under a cosine schedule, momentum 0.9, weight decay 0.001, 5-epoch warmup) with the standard Ultralytics loss gains (box 7.5, cls 0.5, DFL 1.5). Training was deterministic with mixed precision disabled, and only the seed was varied. Augmentation was configured for grayscale radargrams: HSV jitter, mixup, copy-paste, and random erasing were disabled, whereas geometric augmentation comprised mosaic (probability 0.6, disabled for the final 45 epochs), rotation ±6°, translation 0.1, scale 0.35, horizontal flip 0.5, and vertical flip 0.1. Training and latency measurements were performed on a single NVIDIA GeForce RTX 4090 GPU.

## IV. RESULTS

All accuracy values were computed on unaugmented field test images from the Bangkok subsoil surveys alone, that is, on radargrams acquired over high-water-content soft marine clay, with faint and ambiguous expert-identified cavities retained. Absolute mAP50 values are therefore expected to lie well below the 0.87–0.96 range reported for single-source or augmentation-based evaluations; Section V-B accounts for this difference. The comparisons of primary interest are the relative ones between configurations evaluated under identical conditions.

### *A. Performance of the proposed model*

TriView-YOLO achieved an mAP50 of 0.558 ± 0.028 over three seeds at 23.6 GFLOPs and 3.1 ± 0.3 ms per image, the highest accuracy of any configuration in Table II. Per-seed confusion matrices on the 133 annotated test cavities give 76/54/57, 88/66/45, and 87/61/46 true positives, false positives, and false negatives for seeds 0–2, corresponding to precision 0.581 ± 0.009, recall 0.629 ± 0.050, and F1 0.603 ± 0.022 at the default operating point.

### *B. Ablations*

TABLE II
PROPOSED MODEL AND COMPLETE ABLATION RESULTS, BANGKOK FIELD-ONLY TEST SET (MEAN ± SD, THREE SEEDS).

| Model | Precision | Recall | F1-score | mAP50 | CV (%) | GFLOPs |
|---|---|---|---|---|---|---|
| *Proposed model* | | | | | | |
| **MV small (proposed)** [a] | **0.581 ± 0.009** | **0.629 ± 0.050** | **0.603 ± 0.022** | **0.558 ± 0.028** | 4.93 | 23.6 |
| *Single-view (2D), field-only training* | | | | | | |
| 2D nano (random) | 0.522 ± 0.058 | 0.469 ± 0.031 | 0.493 ± 0.027 | 0.497 ± 0.034 | 6.84 | 6.3 |
| 2D small (random) | 0.543 ± 0.075 | 0.469 ± 0.028 | 0.501 ± 0.036 | 0.481 ± 0.043 | 8.94 | 21.2 |
| 2D medium (random) | 0.483 ± 0.031 | 0.452 ± 0.008 | 0.466 ± 0.013 | 0.443 ± 0.016 | 3.61 | 67.1 |
| 2D nano (COCO) | 0.481 ± 0.050 | 0.484 ± 0.055 | 0.478 ± 0.048 | 0.450 ± 0.023 | 5.11 | 6.3 |
| 2D small (COCO) | 0.519 ± 0.013 | 0.439 ± 0.027 | 0.476 ± 0.021 | 0.460 ± 0.018 | 3.91 | 21.2 |
| 2D medium (COCO) | 0.484 ± 0.006 | 0.438 ± 0.036 | 0.459 ± 0.017 | 0.454 ± 0.040 | 8.81 | 67.1 |
| *Single-view (2D), + Roboflow + gprMax* | | | | | | |
| 2D nano (random) | 0.464 ± 0.009 | 0.453 ± 0.018 | 0.457 ± 0.022 | 0.441 ± 0.010 | 2.27 | 6.3 |
| 2D small (random) | 0.440 ± 0.036 | 0.421 ± 0.004 | 0.430 ± 0.016 | 0.424 ± 0.013 | 3.07 | 21.2 |
| 2D medium (random) | 0.509 ± 0.039 | 0.380 ± 0.021 | 0.435 ± 0.008 | 0.418 ± 0.008 | 1.91 | 67.1 |
| 2D nano (COCO) | 0.479 ± 0.072 | 0.423 ± 0.042 | 0.453 ± 0.005 | 0.442 ± 0.012 | 2.71 | 6.3 |
| 2D small (COCO) | 0.476 ± 0.007 | 0.461 ± 0.017 | 0.468 ± 0.012 | 0.424 ± 0.014 | 3.30 | 21.2 |
| 2D medium (COCO) | 0.467 ± 0.074 | 0.408 ± 0.041 | 0.431 ± 0.017 | 0.423 ± 0.039 | 9.22 | 67.1 |
| *Multi-view (MV), field-only training* | | | | | | |
| MV medium | 0.554 ± 0.022 | 0.533 ± 0.012 | *0.543 ± 0.004* | 0.513 ± 0.012 | 2.34 | 63.5 |
| MV small + DINO-P4 | 0.475 ± 0.015 | 0.503 ± 0.031 | 0.488 ± 0.020 | 0.483 ± 0.014 | 2.90 | 58.7 |
| MV medium + DINO-P4 | 0.503 ± 0.036 | 0.466 ± 0.020 | 0.484 ± 0.024 | 0.434 ± 0.032 | 7.37 | 99.9 |
| MV small + DINO-P5 | 0.520 ± 0.055 | *0.556 ± 0.024* | 0.535 ± 0.031 | *0.519 ± 0.006* | 1.16 | 58.2 |
| MV medium + DINO-P5 | *0.571 ± 0.018* | 0.501 ± 0.053 | 0.532 ± 0.023 | 0.518 ± 0.017 | 3.28 | 105.5 |

[a] From the per-seed confusion matrices at the default operating point; ablation rows report validation-curve values at the maximum-F1 point. mAP50 is operating-point-independent and comparable across all rows.

Table II reports the proposed model together with the seventeen ablation configurations, all trained with the same seeds and hyperparameters and evaluated with the same metrics under the fixed splits and the common annotation set of Section III-C. Each ablation row removes or replaces one design element of the proposed model. The proposed model and the eleven field-only ablation rows share the field test split, whereas the six multi-source rows are evaluated on the field-only test images of the multi-source split, so the two training-data conditions are compared across their respective field-only test sets. Within the ablation block, *italics* mark the best ablation value; no ablation reaches the mAP50 of the proposed model. CV is the coefficient of variation of mAP50; GFLOPs are architecture-determined and therefore shared by the two training-data conditions. Inference latency is not tabulated but is plotted for the field-only configurations in Fig. 4.

**Input views.** Removing the two auxiliary views reduces the model to a standard YOLOv12 operating on longitudinal B-scans alone. The best single-view model, 2D nano with random initialization, reached mAP50 0.497 with recall 0.469, whereas the proposed multi-view model reached 0.558 with recall 0.629, and every multi-view ablation without mid-level DINOv3 injection reached mAP50 0.513–0.519. With three seeds these differences cannot be formally significance-tested; they are, however, consistent in direction across seeds and metrics, and the multi-view input is the only factor tested whose removal degraded every accuracy metric. The single-view group also covers scale and initialization: COCO pretraining did not consistently outperform random initialization, consistent with the difference between radargram and natural-image statistics, and larger capacity did not improve accuracy, indicating that dataset size rather than model expressiveness limits single-view performance. All mAP50 CV values were below 10%.

**Training data.** Adding Roboflow Universe and gprMax images to the training set, with field-only validation and a Bangkok-only test set, lowered performance for every single-view configuration in which it was tested. The best multi-source model reached mAP50 0.442, compared with 0.497 for field-only training, although seed-to-seed variation decreased (CV as low as 1.91%). The degradation is consistent with distribution shift: public and synthetic images exhibit clearer, less noisy cavity signatures than field radargrams, which biases the decision boundary away from real survey statistics. This cautions against the common practice of supplementing scarce field GPR data with cleaner auxiliary data and qualifies earlier positive findings on simulation mixing [20], [23].

**DINOv3 enhancement and model scale.** No DINOv3 or larger-scale ablation approached the proposed model. The strongest, MV small + DINO-P5, reached mAP50 0.519 and recall 0.556, against 0.558 and 0.629 for the proposed model at 2.5 times lower computational cost, and MV medium + DINO-P5 achieved the best ablation precision (0.571) but no improvement in mAP50. All plain and DINO-P5 variants exceeded every single-view configuration in mAP50, whereas DINO-P4 underperformed, reaching values as low as 0.434, and was the least stable configuration (CV 7.37%). This contrasts with the gains reported by DINO-YOLO for the same integration concept on natural civil engineering imagery [28]: DINOv3 was pretrained on natural images, whereas radargrams contain reflections, speckle-like noise, and hyperbolic responses, so its features did not transfer under the configurations tested, and mid-level (P4) injection disrupted the learned feature hierarchy. Scaling from small to medium did not improve mAP50 in either group.

### *C. Accuracy–efficiency trade-off*

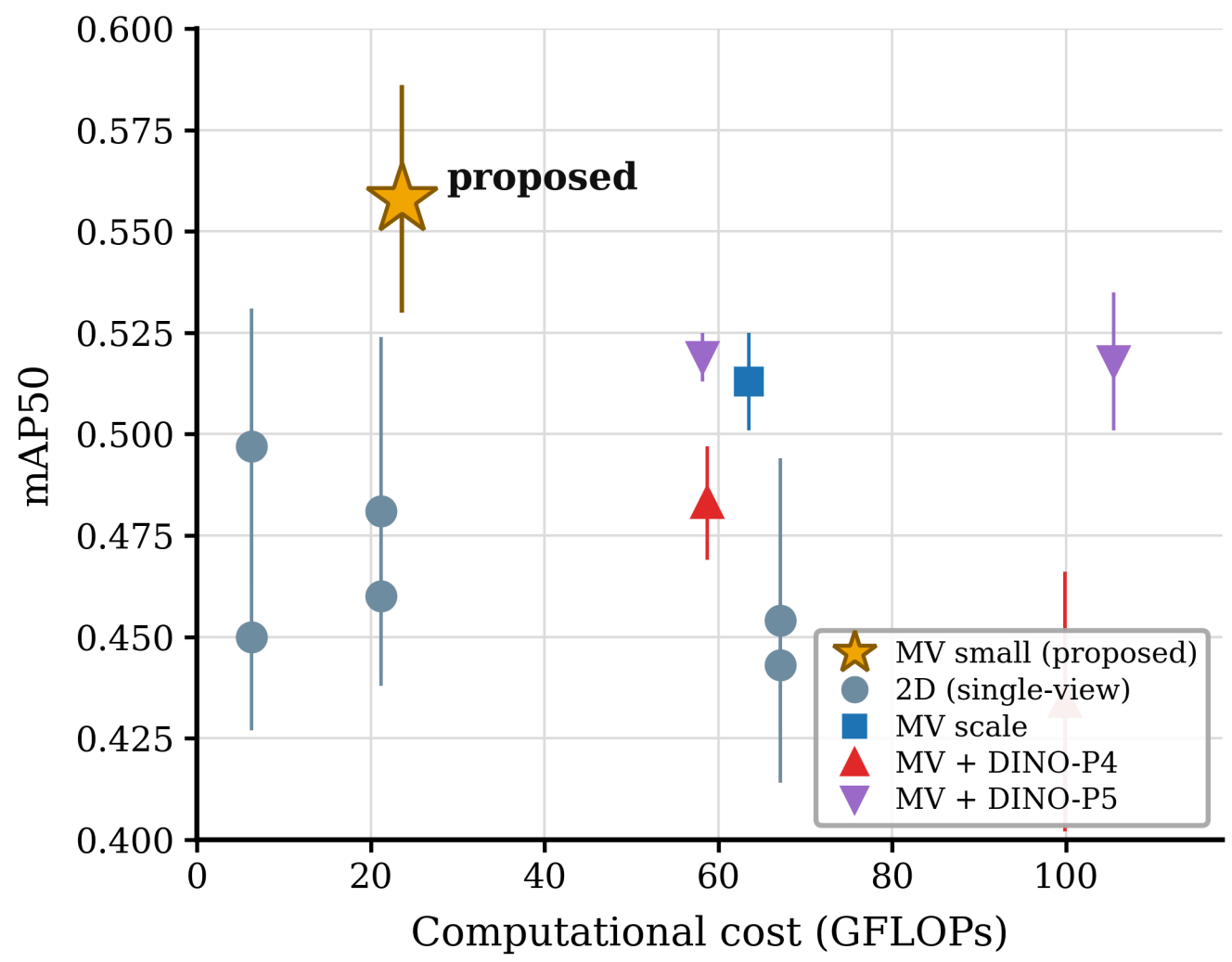

Fig. 3. Accuracy–efficiency trade-off of the proposed model (star) and its field-only ablations: mAP50 versus computational cost. Error bars denote ± SD over three seeds; individual configurations are identified in Table II.

Figures 3 and 4 plot mAP50 against computational cost and against single-image inference latency respectively, for the proposed model (star) and the field-only ablations of Table II; the six multi-source configurations are omitted because each is less accurate than its field-only counterpart at identical cost. Among single-view models, 2D nano with random initialization achieved the highest mAP50 at the lowest cost of the entire set (0.497 at 6.3 GFLOPs and 2.0 ± 0.2 ms). Among multi-view models, the proposed MV small achieved the highest mAP50 at the lowest cost within its group (23.6 GFLOPs, 3.1 ± 0.3 ms), whereas the DINO-P5 variants achieved lower accuracy at 2.5–4.5 times the FLOPs (58.2 and 105.5 GFLOPs) and about twice the latency (6.1 ± 0.1 and 6.8 ± 1.4 ms), and MV medium + DINO-P4 combined the second-highest FLOPs (99.9 GFLOPs) and the highest latency (7.1 ± 0.4 ms) with the lowest accuracy. The accuracy–cost Pareto front in Fig. 3 therefore reduces to two configurations, the 2D nano baseline and the proposed MV small, with every remaining configuration at least as costly as one of them while being less accurate; Fig. 4 gives the same ordering in latency, the only exception being 2D nano with COCO initialization, nominally the fastest model at 1.8 ± 0.2 ms but the less accurate member of the nano pair. Relative to the single-view baseline, the proposed model gains 0.061 mAP50 and 0.160 recall for an increase in latency from 2.0 to 3.1 ms, which remains compatible with field screening throughput.

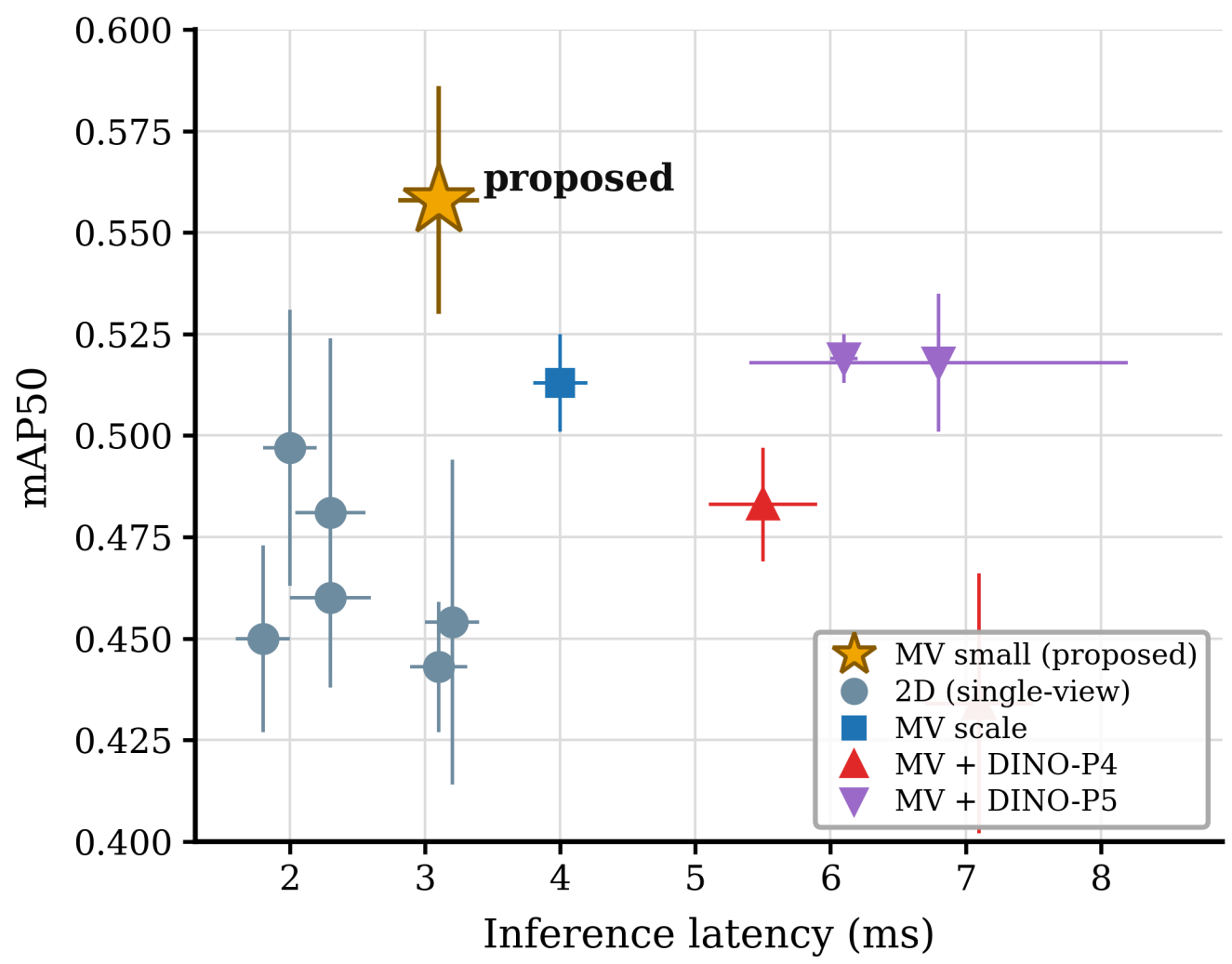

Fig. 4. Accuracy–efficiency trade-off: mAP50 versus single-image inference latency, with latency SD shown horizontally. Symbols and error bars as in Fig. 3.

## V. DISCUSSION

### A. Why multi-view fusion helps in high-attenuation ground

That fusing three co-registered views lifts recall and mAP50 above every single-view configuration at a latency increase of only 1.1 ms admits a physical interpretation beyond the generic benefit of additional input data. In high-water-content marine clay, conductive losses attenuate the radar signal at tens to hundreds of dB/m [30], so the hyperbolic signature of a cavity in a single longitudinal B-scan is frequently degraded to the point of being statistically indistinguishable from clutter; this is the regime in which the single-view baseline missed approximately half of the cavities. The C-scan and the orthogonal cross-section supply evidence that is conditionally independent of the longitudinal view given the cavity, an amplitude anomaly in plan view and a second, orthogonally oriented reflection section, so a cavity whose longitudinal hyperbola falls below the reliable-detection threshold can still be confirmed by the remaining two views. For screening in attenuating ground, where a missed cavity carries a higher cost than a false alarm, this recall-oriented shift is the operationally relevant property.

Three independent negative results point to a common cause. Adding cleaner public and synthetic B-scans lowered field mAP50 from 0.497 to 0.442; COCO-pretrained initialization did not consistently outperform random initialization; and DINOv3 enhancement brought no consistent gain, with mid-level injection proving harmful. In all three cases, features or data from cleaner or natural-image domains were transferred to a problem whose inputs are speckle-laden, attenuation-degraded field radargrams. The consistency across three transfer mechanisms supports a stronger conclusion than any single experiment would permit: wet-ground field GPR statistics differ sufficiently from both synthetic GPR and natural images that direct transfer displaces the decision boundary unfavorably. This is consistent with the domain-shift sensitivity reported for a state-of-the-art GPR detector under zero-shot evaluation [25]. The constructive implication is that the requirement is not for more auxiliary data but for matched auxiliary data, such as simulations of conductive, high-moisture soils with realistic clutter, or self-supervised pretraining on large unlabeled corpora of real radargrams, for which the multi-view input provides a natural pretext structure.

### B. Why the absolute mAP50 is lower than in previous studies

The best mAP50 obtained here (0.558) is markedly lower than the range of 0.87–0.96 reported in previous GPR defect detection studies (Table I), a difference attributable to ground conditions and evaluation protocol rather than to a deficiency of the detector. All previously published cavity detectors in Table I were developed and evaluated on firm or well-drained ground, whereas the present study is evaluated entirely on soft, high-water-content marine clay, in which attenuation exceeds that of dry sand by up to four orders of magnitude [30] and target visibility decreases with increasing moisture [31]. Test-set composition compounds this effect: several of the highest reported accuracies were obtained on data expanded eightfold by augmentation before splitting [16], on pools in which synthetic B-scans appear in the evaluation itself [23], or on datasets retaining only images pre-selected as containing typical cavity signatures [21], whereas the present test set is unaugmented Bangkok field imagery that keeps faint and ambiguous expert-identified cavities, reported as means over three seeds.

Within the GPR literature, detectors evaluated on field data alone score well below the headline range: recall 0.587 on a 209-km highway survey [22], YOLOv8 recall 0.663 on a street utility survey [24], and a recent state-of-the-art detector reaching mAP50 0.696 on its own benchmark but 0.558, essentially the level reported here, under zero-shot evaluation [25]. The most directly comparable study, covering 1,250 km of urban 3D GPR partly over water-rich and soft soils, reported single-view void precision and recall of approximately 0.32, with 63% of detections at fifteen unseen sites proving false [26]. The figures reported here are therefore best interpreted as a reproducible baseline for cavity screening in adverse soft ground, against which future domain-adapted methods can be compared.

### C. Practical implications, transferability, and limitations

For road agencies in soft-ground deltaic cities, three practical points follow. Where co-registered C-scan and cross-section views can be exported from the survey software, which is the standard output of the multichannel systems already used for cavity screening, the multi-view model is preferable, because its benefit is concentrated in the weak-signature cases that dominate wet-clay surveys; where only longitudinal B-scans are available, the 2D nano model at 6.3 GFLOPs remains a viable low-cost option. The operating threshold should be set by the risk asymmetry of the application rather than by maximizing F1, since for screening in advance of pavement failure the cost of a missed cavity justifies the false-alarm rate at the recall-oriented operating point; at this point roughly 40% of flagged locations prove to be false alarms (54–66 false positives against 76–88 true positives per seed), a verification workload not yet quantified in an operational study. Finally, the reported accuracy level indicates what an agency should expect from automated screening in this environment: procurement criteria calibrated against the 0.87–0.96 figures reported for favorable-ground studies would reject systems performing at the realistic state of the art.

The TripleInputConv design modifies only the first layer and requires labels on one primary view, so the mechanism should transfer to other problems providing complementary co-registered views, such as bridge-deck assessment combining GPR with thermography, tunnel-lining and utility inspection, or multispectral road-scene sensing [33]. This remains a hypothesis, since no experiment outside GPR is reported here. Two boundary conditions qualify the transfer: when views are unregistered or geometrically misaligned, input-level concatenation loses its advantage and later fusion with learned correspondence is preferable, and when the statistical gap between modalities is large, halfway fusion with cross-modal interaction can outperform simple early fusion [33]. The released implementation [37] accepts arbitrary co-registered image triplets.

Several limitations qualify the results. The dataset is of modest size (1,600 field samples); evaluation on fully independent routes was not performed, so the accuracies measure within-survey generalization; accuracy on the Japanese routes was not measured separately; and expert positions were not exhaustively verified by excavation. The three-seed protocol limits statistical power, accuracy is reported at IoU 0.5 only, and no previously published detector was retrained on this dataset, so Table I provides context rather than a controlled comparison. Neither the direct 9-channel early-fusion stem nor a one-channel-per-view variant was trained as a control, so the preference for branch-then-fuse rests on the initialization and cost arguments rather than on measurement. The Bangkok environment imposes four further constraints: usable penetration below approximately 2 m means that deeper voids are absent from the radargrams [18]; water-filled cavities below the water table produce weak or absent reflections and are therefore under-represented; annotation certainty is lowest where the environment is most adverse, so part of the residual error may reflect ground-truth rather than model error; and the moisture regime was sampled at a single point in time, leaving seasonal robustness untested. Because the high-attenuation evidence originates from a single deltaic marine-clay setting, the results support screening in marine clay comparable to that of Bangkok rather than in soft ground generally.

## VI. CONCLUSIONS AND FUTURE WORK

This study proposed TriView-YOLO, a multi-view YOLOv12 detector for subsurface road cavity detection from GPR images, in which three co-registered views are fused by a TripleInputConv stem, and isolated the contribution of each design choice through seeded ablations evaluated on a test set drawn exclusively from Bangkok subsoil surveys. The proposed model reaches mAP50 0.558 ± 0.028 at 23.6 GFLOPs and 3.1 ms per image, the highest accuracy of any configuration evaluated; removing the auxiliary views reduced mAP50 to at most 0.497 ± 0.034 and recall from 0.629 to 0.469, with the differences consistent in direction across seeds although the three-seed protocol precludes formal significance testing. Every accuracy reported here was measured over 8–15 m of soft marine clay with 80–140% water content, where strong attenuation weakens cavity reflections and interpretation has previously been manual, so these figures provide a realistic baseline for automated screening in soft-ground deltaic cities. Adding public and synthetic B-scans reduced field-evaluated mAP50 to 0.442 ± 0.012, and neither COCO initialization nor DINOv3 enhancement consistently improved accuracy, which identifies GPR-specific pretraining rather than natural-image transfer as the promising direction. The selected models were implemented in a web application with seed-ensemble inference; because roughly 40% of flagged locations are false alarms at the deployed operating point and false negatives remain, outputs must be treated as preliminary results subject to expert verification.

The highest-priority extension is to enlarge the field dataset with surveys on fully independent routes, so that generalization can be measured across rather than within survey campaigns, together with repeated surveys across wet and dry seasons, lower-frequency antennas to extend the screening envelope beyond the current 1–2 m limit, systematic excavation verification, and blind re-annotation with inter-rater statistics. Because cleaner synthetic data degraded field performance, simulation research should target realism rather than volume, through models of conductive, high-moisture soils with realistic clutter and domain-adaptation training that treats synthetic and field data as distinct domains. On the model side, subsequent steps include training the direct 9-channel and one-channel-per-view stems as controls, expanding the seed count to support formal significance testing and reporting mAP50–95, retraining representative prior detectors on this dataset for controlled comparison, benchmarking newer YOLO generations, and replacing the fixed 1 × 1 fusion with learned per-view gating so that a corrupted or missing view can be down-weighted.

## ACKNOWLEDGMENTS

The authors thank the Department of Civil Engineering, King Mongkut's University of Technology Thonburi, for funding support, and Ehime University and Canaan Geo Research Co., Ltd. for field investigation support and research exchange during the internship in Japan.

## DATA AVAILABILITY

The source code of the multi-view (triple-input) YOLOv12 architecture is available at https://github.com/Sompote/Triple-dino-yolov12 [37], and the trained models are demonstrated at https://huggingface.co/spaces/QSuphawut/3d-cavity-nodino. Field GPR data are available from the corresponding author upon reasonable request, subject to survey-owner permission.